\documentclass[conference]{IEEEtran}
\IEEEoverridecommandlockouts
\usepackage{cite}
\usepackage{amsmath,amssymb,amsfonts}
\usepackage{algorithmic}
\usepackage{graphicx}
\usepackage{textcomp}
\usepackage{xcolor}
\usepackage{tabularx}
\usepackage{booktabs}

\def\BibTeX{{\rm B\kern-.05em{\sc i\kern-.025em b}\kern-.08em
    T\kern-.1667em\lower.7ex\hbox{E}\kern-.125emX}}
\begin{document}

\title{Bridging Language Gaps with Adaptive RAG: Improving Indonesian Language Question Answering\\
}

\author{\IEEEauthorblockN{1\textsuperscript{st} William Christian}
\IEEEauthorblockA{
\textit{Computer Science Department} \\
\textit{School of Computer Science} \\
\textit{Bina Nusantara University} \\
Jakarta, Indonesia \\
william.christian001@binus.ac.id}

\and
\IEEEauthorblockN{2\textsuperscript{nd} Daniel Adamlu}
\IEEEauthorblockA{
\textit{Computer Science Department} \\
\textit{School of Computer Science} \\
\textit{Bina Nusantara University} \\
Jakarta, Indonesia \\
daniel.adamlu@binus.ac.id}

\and

\IEEEauthorblockN{3\textsuperscript{rd} Adrian Yu}
\IEEEauthorblockA{
\textit{Computer Science Department} \\
\textit{School of Computer Science} \\
\textit{Bina Nusantara University} \\
Jakarta, Indonesia \\
adrian.yu@binus.ac.id}

\and

\IEEEauthorblockN{4\textsuperscript{th} Derwin Suhartono}
\IEEEauthorblockA{
\textit{Computer Science Department} \\
\textit{School of Computer Science} \\
\textit{Bina Nusantara University} \\
Jakarta, Indonesia \\
dsuhartono@binus.edu}
}

\maketitle

\begin{abstract}
Question Answering (QA) has seen significant improvements with the advancement of machine learning models, further studies enhanced this question answering system by retrieving external information, called Retrieval-Augmented Generation (RAG) to produce more accurate and informative answers. However, these state-of-the-art-performance is predominantly in English language. To address this gap we made an effort of bridging language gaps by incorporating Adaptive RAG system to Indonesian language. Adaptive RAG system integrates a classifier whose task is to distinguish the question complexity, which in turn determines the strategy for answering the question. To overcome the limited availability of Indonesian language dataset, our study employs machine translation as data augmentation approach. Experiments show reliable question complexity classifier; however, we observed significant inconsistencies in multi-retrieval answering strategy which negatively impacted the overall evaluation when this strategy was applied. These findings highlight both the promise and challenges of question answering in low-resource language suggesting directions for future improvement.
\end{abstract}


\section{Introduction}\label{sec:introduction}

Recent Large Language Models (LLMs) have shown incredible performance for a lot of Natural Language tasks. Commercial models such as DeepSeek, Gemini, Claude, and GPT \cite{deepseek} \cite{gemini} \cite{claude} \cite{gpt}, or even an open-source model such as Qwen and Gemma \cite{qwen} \cite{gemma} show state-of-the-art performance in tasks such as language understanding, reasoning, mathematics, coding, including question answering (QA).

However, despite the advancement of LLMs in all tasks in natural language processing, they still have problems answering questions that require a knowledge-intensive background, often resulting in hallucination answers \cite{hallu_llm}. LLMs often provide accurate answers when entities mentioned in the question are present in their training data. Furthermore, the performance of the models has a significant correlation with the entity popularity; less popular entities are often not answered accurately by LLMs \cite{trust_llm}. Updating the LLM's knowledge frequently is not a good solution since the training of LLM with billions or even trillions of data from all over the internet takes too much time.

In contrast, recent studies have demonstrated that augmenting non-parametric knowledge (information not contained in the model's training data) to the question-answering method commonly referred to as Retrieval Augmented Generation (RAG) \cite{first_rag}, even smaller models outperform larger models in terms of parameters \cite{atlas_rag}. RAG operates by introducing an additional retrieval layer prior to the input of the LLM. This layer retrieves relevant information from internal or external sources, such as knowledge bases or the internet, and incorporates it into the model’s input. One of the most impacted tasks is question answering in many fields in real-world application such as medicine, finance, and healthcare \cite{medicine_rag} \cite{financial_rag} \cite{medical_rag}. In the beginning, RAG systems were designed simply using single-step extraction, where the system retrieves the information only once and then augments the information to the model's input \cite{singlestep_first} \cite{singlestep_second}. This method usually works fine with non-complex questions, where only one piece of information was needed to provide sufficient context for the LLM to generate an accurate response. Nevertheless, recent studies focus on more advanced types of retrieval process, such as implementing chain-of-thought method in retrieval process, which shows significant performance improvements on multiple datasets and methods \cite{ircot}.  It is also important to note that not all questions require external retrieval.  When the information is already contained in the LLM's training data and pertains to relatively popular entities, the model is often capable of answering correctly without retrieval or with only a single retrieval step \cite{adaptive_rag}.

Another challenge that arises from the findings of different types of methods is the significant time and computational costs required to run these systems, especially multi-retrieval systems, where in each iteration the retrieval process queries the LLM for the answer while appending the previous context \cite{ircot}. As a result, running such systems can consume substantial computational resources when the LLM is deployed locally or incur high costs when relying on outsourced APIs, in addition to increasing the latency of response generation. These drawbacks become even more pronounced in cases where the LLM does not actually require external information to answer the question. 

To address the limitations of LLMs in question answering, multiple studies have proposed different forms of strategic retrieval, often referred to as adaptive retrieval. In this approach, the retrieval process is driven by the complexity of the question, allowing the system to save time and computational costs when answering non-complex questions \cite{trust_llm} \cite{adaptive_rag} \cite{vary_reasoning}. Despite these advances, a major gap remains in the context of low-resource languages. All three of the aforementioned studies on Adaptive RAG systems have been conducted exclusively in English. This is unsurprising, given the dominance of English in research and technology. Although English is not the world's most widely spoken native language, it is the second most common language \cite{english_dominance}, which has strongly skewed the development of LLMs toward English.

Another challenge lies in the availability of high-quality datasets for the question-answering tasks in low-resource languages. While several datasets exist for Indonesian language QA \cite{indoqa} \cite{qasina}, none have been designed specifically for use in multi-retrieval settings. This lack of resources significantly limits the exploration of adaptive RAG in Indonesian language. In response to these challenges, this study contributes the following:  
\begin{itemize}
    \item The creation of a multi-retrieval dataset for Indonesian language. We translated existing English benchmark datasets into Indonesian language, enabling the systematic evaluation of multi-retrieval question answering in a low-resource languages.
    \item Exploration of adaptive RAG in Indonesian language. We evaluate the performance of adaptive retrieval-augmented generation methods for Indonesian language question answering tasks.
    \item Analysis of LLM challenges in a low-resource language. We highlight the limitations of LLMs in applying them to Indonesian language, particularly in terms of accuracy, efficiency, and retrieval dependence.
    \item Incorporating methods from Adaptive-RAG \cite{adaptive_rag} into the Indonesian language context.
\end{itemize}

The remainder of this paper is organized as follows. Section~\ref{sec:related} reviews related works, including prior research on Indonesian language question answering datasets, language translation efforts, and studies on adaptive retrieval. Section~\ref{sec:methods} describes the methods used in this study, providing details on the approaches and techniques applied. Section~\ref{sec:experiments} outlines the experimental setup and explains how the proposed methods were implemented. Section~\ref{sec:results} presents the results of the experiments, followed by a discussion and interpretation of the findings. Finally, Section~\ref{sec:conclusion} concludes the paper and highlights potential directions for future research.

\section{Related Works}
\label{sec:related}

\subsection{Text Classification}
Text classification has long been a fundamental task in Natural Language Processing, and its importance has grown significantly with the rise of digital content. In today’s internet era, the vast amount of textual data being generated makes text classification essential for efficiently organizing, managing, and extracting meaningful information from large-scale text corpora. Early approaches to text classification primarily relied on classical machine learning techniques such as Decision Trees, Naive Bayes, Rule Induction, Nearest Neighbor, and Support Vector Machines \cite{classic_tc}. During that period, text classification remained a major area of research, with efforts focused on designing effective feature representations and optimizing these algorithms for various text domains. Currently, the most widely adopted models for text classification—and many other machine learning tasks—are based on the Transformer architecture, which has demonstrated exceptional performance across a broad range of applications \cite{transformer}. 

The most popular models used for text classification tasks are transformer-based models such as BERT \cite{bert}. Many subsequent models build upon BERT as a base model, focusing on improving and optimizing its performance. For example, RoBERTa \cite{roberta} enhances BERT through more robust training strategies, while DistilBERT \cite{distilbert} reduces the model size while maintaining much of BERT’s performance. Other models aim to address multilingual tasks, such as XLM-RoBERTa \cite{xlm-roberta}. Leveraging the power of pretrained BERT, numerous efforts have also been made to fine-tune the base model for specific languages. In our study, we focus on IndoBERT \cite{indobert}, a BERT variant pretrained for the Indonesian language. We have also conducted a study using IndoBERT to improve the performance of emotion classification. Our findings show that removing stop words decreases the model’s ability to understand sentence meaning, leading to improvements across several evaluation metrics, including accuracy, precision, recall, and F1-score after skipping stop words removal \cite{emotion_indobert}.

\subsection{Question Answering}
Question Answering is a task in Natural Language Processing (NLP) that aims to automatically answer precisely to a question. In the beginning, the method for answering questions is by performing Information Retrieval (IR), which returns ranked documents or passages based on the similarity of sentences, with additional process such as question analysis, question classification, information filtering and information ordering  \cite{qa_definition}, the main objective of the the QA task is to correctly answer a question. On another hand, Open Domain QA is an another task that is specifically focused on trying to answer question where the answer is not in the parametric knowledge \cite{realm}. 

To develop and evaluate such systems, there must exist a way to benchmark each system for evaluating the performance of each model, that's where Question Answering dataset becomes crucial in benchmarking \cite{system_verification}. These datasets consists of carefully constructed pairs of questions and corresponding answer, in some cases additional columns exist to help give more context to each row such as information, question type, and even some dataset serve give multiple answers \cite{musique} \cite{popqa} \cite{squad}. There are a lot of dataset already created in the task of Question Answering especially for the English language with different specific task for single retrieval \cite{squad} and multi retrieval \cite{musique} \cite{popqa} \cite{hotpotqa}.

Unfortunately, Indonesian language QA dataset remain scarce and often limited in quality,  making it challenging to conduct research in this area \cite{indoqa} \cite{qasina}. In particular, dataset that is specified for doing multi-step retrieval is not available represents a significant obstacle, one that we aim to address in the following chapter.

\subsection{Machine Translation}
Machine translation has been one of the most research areas in Natural Language Processing where the main objective of the task is to transfer a sentence from the source language to the target language while preserving the entirety of the meaning from the original sentence \cite{mt-definition} \cite{mt-progress}. Early attempts to address this issue relied on Rule-Based Machine Translation (RBMT) \cite{rbmt}, which focused on constructing predefined linguistic and grammatical rules to translate from a source language to a target language. However, the main limitation of this approach lies in the highly time-consuming process of manually creating rules to cover the vast vocabulary and structures of both languages. After the popularity of RBMT, a new approach known as Statistical Machine Translation (SMT) was introduced \cite{smt}. This method relies on probabilistic models of language, leveraging large bilingual corpora to estimate the likelihood of translations. Unlike RBMT, SMT is more effective in generating outputs for sentences that are not explicitly covered by predefined grammatical rules, as it learns translation patterns directly from data. However, following the development of Statistical Machine Translation, the field of Natural Language Processing saw limited breakthroughs in machine translation until the advent of Neural Networks, which led to the emergence of Neural Machine Translation (NMT) \cite{nmt}. Numerous studies have since demonstrated that NMT achieves state-of-the-art performance across a wide range of translation tasks.

With the rise of Neural Machine Translation, the dataset translation in our study was carried out using OPUS-MT \cite{opus-mt}, a model based on Marian \cite{marian} and developed by the University of Helsinki. The details of the translation process are further explained in Section~\ref{sec:methods}.

\subsection{Single and Multi Step Retrieval}
Single-step retrieval is the process in which a RAG system performs only one round of retrieval. The information sources can include vector databases, existing datasets, the internet, Wikipedia, and many others \cite{multifc} \cite{internet_augmented} \cite{trillion}. Although the process may appear simple—finding the most relevant information for a given query—it is one of the most challenging components of RAG systems. Various retrieval methods have been developed to address this, such as lexical matching, which relies on exact word overlap between the query and documents, with common techniques including BM25 \cite{bm25} and TF-IDF \cite{tfidf}. Another widely used approach is dense retrieval, which, while similar to lexical matching in aiming to find similarity, instead represents text as embeddings for richer semantic representation. More recently, Graph RAG has emerged as a popular method, where documents or information are structured into a graph to capture relationships using LLMs \cite{graph_rag}. Ultimately, regardless of the method, the goal remains the same: to retrieve the most relevant information for the query.

However recent studies around retrieval methods has been focused in multi-step retrieval, improving the ability of retrieval system to gather multiple information that is needed for answering specifically complex question \cite{ircot} \cite{adaptive_rag}. Studies have proposed methods such as SelfAsk \cite{self-ask} prompts an LLM to breakdown question into sub-questions so the prompt can get better by knowing what questions should be asked to get the final answer. Similarly DecomP or decomposed prompting \cite{decompose} uses LLM to breakdown complex question into a more simpler task that is more manageable to complete. With the introduction of chain-of-thought prompting—where a model is encouraged to explain its reasoning process step by step rather than providing an answer directly—the reasoning performance of LLMs has improved significantly \cite{chain-of-thought}. Building on this idea, several approaches have been proposed to enhance retrieval and reasoning. For example, ReAct \cite{re-act} encourages LLMs to interleave reasoning with actions in their outputs, although this method tends to work more effectively with larger or fine-tuned models rather than smaller ones. More recently, Adaptive-RAG \cite{adaptive_rag} integrates the state-of-the-art IRCoT \cite{ircot} for multi-step retrieval, where an interleaving approach guides the retrieval process so that prompts can more effectively provide the context needed for answering questions comprehensively.

\subsection{Large Language Models}
Large Language Models (LLMs) have emerged and become an important aspect in the Natural Language Processing field, demonstrating state-of-the-art performance across a wide variety of task such as text generation, text classification, summarization, reasoning and also question answering which we used in this study. These models, typically built on the Transformer architecture, and are trained on a massive corpora of text and capable of capturing complex linguistic patterns at scale and different languages \cite{gpt-3} \cite{gpt} \cite{deepseek}.

While exclusive LLMs such as GPT-4, Claude, and much more \cite{deepseek} \cite{claude} \cite{gpt} have dominated the landscape, recent years have seen the rise of powerful open-source alternatives, which provide researchers with greater flexibility, transparency, and accessibility. Among the most notable are LLaMA, developed by Meta, which focuses on efficient scaling with strong performance \cite{llama}; Qwen, developed by Alibaba, which emphasizes multilingual and domain adaptability \cite{qwen}; and Gemma, developed by Google, which provides lightweight yet high-performing models designed for research and deployment \cite{gemma}. In this study, we leverage these open-source LLMs for text classification, taking advantage of their adaptability and ease of fine-tuning for specific downstream tasks.

\subsection{Adaptive RAG}
In real-world question answering tasks, questions complexity can vary vastly depending on the information needed for answering those question. Some require multiple pieces of supporting information, others can be answered with a single source, and even certain question doesn't even require any information retrieval process as the necessary knowledge is already contained in within the parametric memory of the LLM. However, several approaches have been proposed in recognizing patterns of question's complexity. One such method adapts retrieval method used for information retrieval based on the entities in the question popularity, since it shows from the study that the more popular the entity the better LLM can answer the question correctly \cite{trust_llm}. Nevertheless, this approach remains limited when handling inherently complex questions, even if the entities involved are popular. Newer adaptive methods have also arise, which focuses in using instruction finetuning with diverse tasks and utilizing chain-of-thought makes language models more flexible and capable of adapting to wide range of question complexity.
Although there has been several proposed idea to build adaptive systems for Retrieval Augmented Generated systems, one of the best performing approaches is Adaptive RAG \cite{adaptive_rag}. This method uses text classifier to determine the complexity level of a given question that in return adjust the retrieval process needed for answering the question. The process starts from dataset labelling based on the question complxity and then using the labelled dataset to train a language model classifier. In this study, we extend and adapt the Adaptive RAG method for the Indonesian language.

\section{Methods}
\label{sec:methods}
In this section, we describe our approach, which involves translating the HotpotQA dataset into a low-resource language, namely Indonesian, followed by the dataset labeling process. The explanation also includes retrieval methods and evaluation metrics. We then present the training of our classifier, designed to classify the complexity of questions, and conclude with an evaluation of the overall system.

\subsection{Dataset Translation}
Unfortunately, due to the scarcity of high-quality Indonesian language multi-hop datasets, and given the importance of evaluating our method on such data, we created our own dataset by translating the HotpotQA dataset into Indonesian language. The translation was carried out entirely using an OPUS-MT translation model \cite{opus-mt} without manual post-editing. While this approach may introduce minor translation artifacts, it allows for efficient large-scale dataset construction while preserving the multi-hop reasoning structure of the original HotpotQA dataset.

\subsection{Retrieval Method}
\begin{figure}[h]
	\centering
	\includegraphics[width=.9\columnwidth]{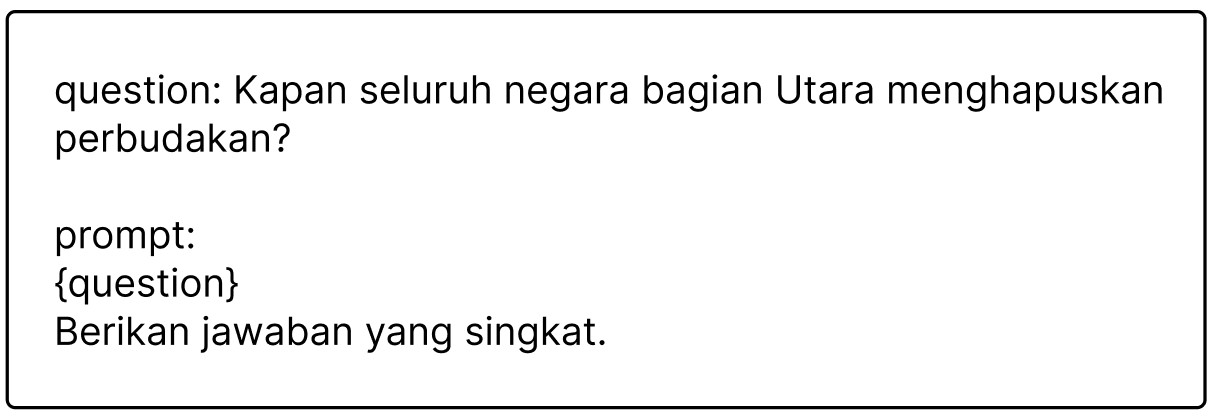}
	\caption{\textbf{Non Retrieval Method.} Formatted prompt for non-retrieval method}
	\label{fig:non_retrieval_graphic}
\end{figure}

\begin{figure}[h]
    \centering
    \includegraphics[width=0.8\linewidth]{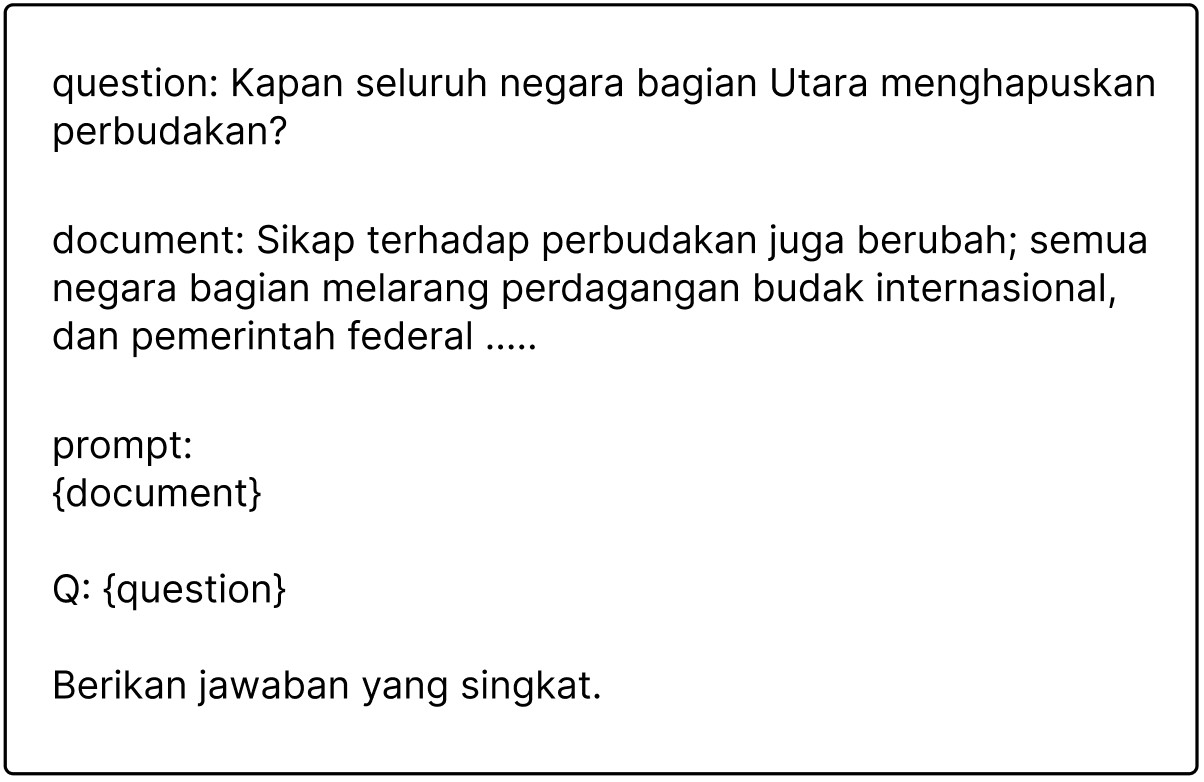}
    \caption{\textbf{Single Retrieval Method.} Formatted prompt for single-retrieval method}
    \label{fig:single_retrieval_graphic}
\end{figure}

\begin{figure}
    \centering
    \includegraphics[width=0.8\linewidth]{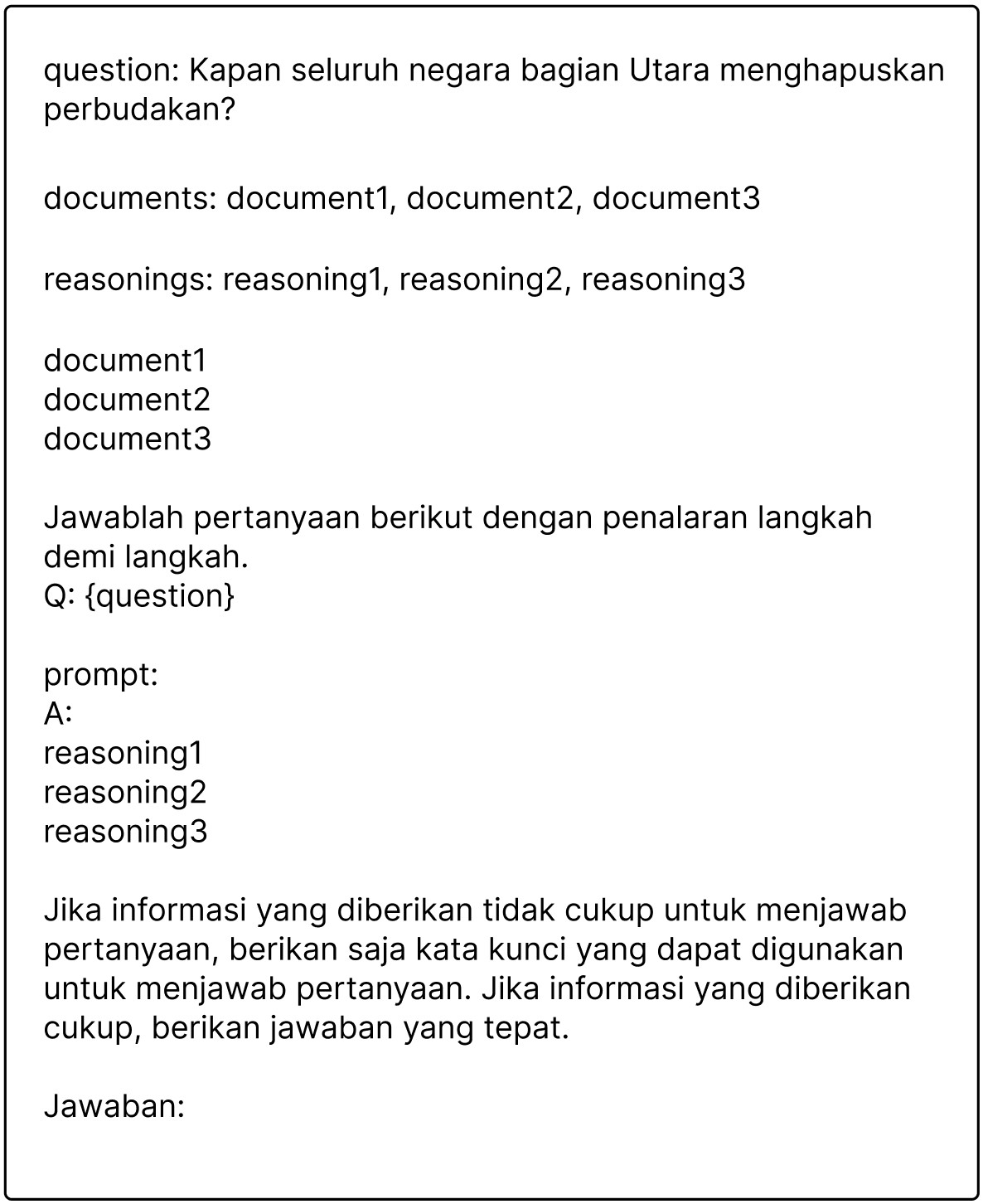}
    \caption{\textbf{Multi Retrieval Method.} Formatted prompt for multi-retrieval method}
    \label{fig:multi_retrieval_graphic}
\end{figure}

Since our paper is inspired by the Adaptive-RAG research study \cite{adaptive_rag}, we adopt a similar concept for the retrieval process but introduce several modifications to improve its ability to follow instructions in the Indonesian language. In this subsection we will explain in detail the implementation of non-retrieval, single-retrieval and multi-retrieval methods. Additionally, the question used in Figures \ref{fig:non_retrieval_graphic}, \ref{fig:single_retrieval_graphic}, and \ref{fig:multi_retrieval_graphic} translates to: “When did all the Northern states abolish slavery?”

\textbf{Non-Retrieval} is the most efficient method in terms of time and computational cost. The method doesn't require to obtain any context for answering the question, dependent to the LLM's parametric knowledge. The prompt used for the non-retrieval process is illustrated in Figure \ref{fig:non_retrieval_graphic}. The instruction "Berikan jawaban yang singkat" (translated as "give a short answer") is included to constrain the LLM’s output. This is necessary because, without such guidance, the model often generates overly detailed explanations, whereas the dataset only requires simple and concise answers.

\textbf{Single Retrieval.} As the name suggests, this method performs the retrieval process only once, utilizing the Elastic Search engine. The input question is first passed to the single retrieval function, which preprocesses it by removing unnecessary symbols such as \texttt{*}, \texttt{!}, and \texttt{.}, as these do not contribute to effective information retrieval. The cleaned question is then used as the query to retrieve relevant documents. After retrieval, the prompt is constructed as illustrated in Figure \ref{fig:single_retrieval_graphic}. The instruction "Berikan jawaban yang singkat" (translated as "give a short answer") is included to guide the LLM to provide concise responses rather than overly detailed explanations.

\textbf{Multi Retrieval.} Our multi-retrieval approach is inspired by the IRCoT (Interleaving Retrieval with Chain-of-Thought) framework \cite{ircot}, which iteratively intertwines retrieval and reasoning to better support multi-step question answering. Unlike the single retrieval method, the question is not only used once for retrieving the information, but the method focuses on utilizing the LLM for guiding the retrieval process. The method is divided into two steps: the retrieve-step and the reason-step.

\begin{itemize}
    \item Retrieve-step: The newly created COT sentence is formulated as a refined query to the retrieval engine (ElasticSearch), which then returns relevant document to support the next cycle.
    \item Reason-step: LLM is used in the current step, where it produces the next COT sentence based on the original question, all previously retrieved documents, and previous reasoning steps. Detailed of the prompt can be seen on Figure \ref{fig:multi_retrieval_graphic}.
\end{itemize}

Each cycle is always started by the retrieve-step, which at the first iteration the original question is used for retrieving the information. The cycle will ends until a predefined termination condition is met, in our study the termination is if there is a keyword "Jawaban" in the reason-step result or if the cycle has happened 5 times. By dynamically guiding the retrieval process, the method increases the likelihood of retrieving all the information necessary to answer the question. 

We used the prompt “Jawablah pertanyaan berikut dengan penalaran langkah demi langkah.” (translated: “Answer the following questions with step-by-step reasoning.”) to make the LLM reason step-by-step and output the key information needed for muli-retrieval method. The prompt “Jika informasi yang diberikan tidak cukup untuk menjawab pertanyaan, berikan saja kata kunci yang dapat digunakan untuk menjawab pertanyaan. Jika informasi yang diberikan cukup, berikan jawaban yang tepat.” (translated as: “If the information provided is insufficient to answer the question, simply provide keywords that can be used to answer the question. If the information provided is sufficient, provide the correct answer.”) was designed to ensure that the LLM outputs important keywords needed for retrieval in the next reasoning step. However, if the given information is sufficient to answer the question, the LLM directly outputs the answer.

\subsection{Evaluation Metrics}
In our study, several aspects need to be evaluated in our study. To assess the prediction of RAG response, the F1 score, the exact match and the accuracy are used as evaluation metrics. F1 score measures the number of overlapping words between the gold (the ground truth from the dataset) and the predicted answer. Exact match, on the other hand, evaluates whether the gold and prediction answer is exactly the same. Accuracy which is closely related to Exact Match, is computed as the ratio of questions with exact matches to the total number of questions. In addition to these, we also include efficiency related metrics for the RAG systems we are experimenting. Time, which measures the duration of the answering process. Lastly Retrieval Count, which records the number of retrieval operations performed for each answer.

\subsection{Data labeling}
Data labeling is a crucial process in our research, as we require a dataset where each question is categorized by its complexity. Using IndoQA and QASiNa dataset mentioned above, we assign a complexity label to every question. Before the labeling process, each question undergoes preprocessing to remove unnecessary characters, ensuring only relevant text is used for the labeling process. Each row from the processed dataset is then subjected to the labeling process. For this step, we apply three answering methods: non-retrieval, single-retrieval and multi-retrieval. Each method generates an answer for the given question, which will also output the evaluated result using exact match and F1 score. These evaluation results are then used to determine the appropriate complexity for each question. The class labels for question complexity are defined as follows: “A” for questions that can be answered without any retrieval process, “B” for questions that require a single retrieval step, and “C” for questions that require multiple retrieval steps to be answered. 
To assign a complexity label, we evaluate each question using the three answering methods in a stepwise manner.
\begin{itemize}
    \item A question is labeled A if the non-retrieval method either achieves an exact match or produces a higher F1-score than the single-retrieval method.
    \item A question is labeled B if the single-retrieval method achieves an exact match or produces a higher F1-score than the multi-retrieval method.
    \item A question is labeled C if the multi-retrieval method achieves an exact match, or if none of the above conditions are met.
\end{itemize}

However for the translated HotpotQA dataset is not subjected to the labeling process described above, since it is naturally designed for the task of multihop question answering. Therefore, all questions in the HotpotQA dataset are directly assigned to the label "C".

\subsection{Question Classifier}
As described in Section \ref{sec:methods}, Adaptive RAG incorporates a text classifier or referred to in this study as question classifier that used for determining the complexity of the questions. For the text classification task, we fine-tuned a pretrained language model, IndoBERT, on our dataset. We utilized labeled data obtained from the annotation process and ensuring distribution of each label in the dataset was balanced. Balancing of the dataset is done using two methods, which is augmentation and undersampling, where least amount of data from the dataset is augmented to become twice as many, and other data label is undersampled to match the size of the least of data. Data augmentation technique that was used are synonym replacement, in which words in a sentence were substituted with alternatives from a curated list of synonym to generate semantically equivalent variants.

We did not apply any additional preprocessing steps, particularly stopword removal, as our previous research shown that this process does not improve the model performance in text classification especially emotion classification and instead removing meaningful information from the sentences \cite{emotion_indobert}. The resulting processed data was then used for fine-tuning, where we adopted the hyperparameters from the best-performing model in our prior study. The evaluation metrics used for fine-tuning the language model were accuracy, precision, recall and F1 score. In addition to this metrics, we also utilizes a confusion matrix to analyze the overall performance across different classes.

\begin{figure*}[h]
    \centering
    \includegraphics[width=.9\textwidth]{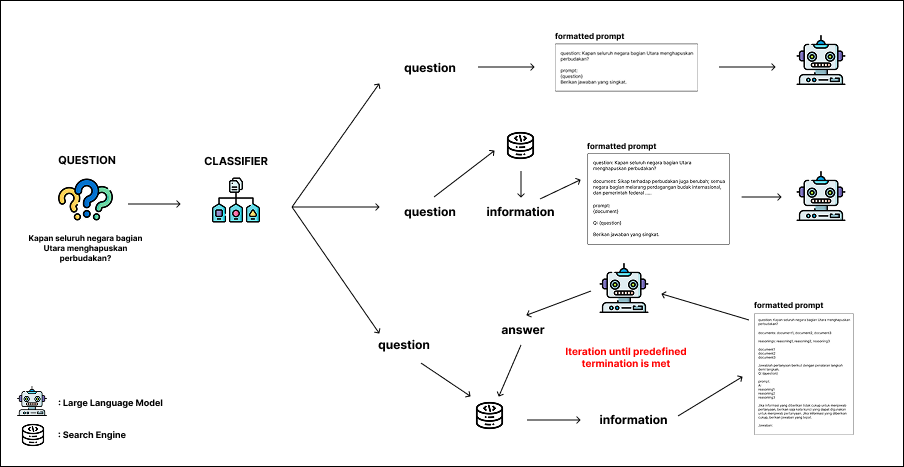}
    \caption{\textbf{Indonesian Language Adaptive RAG system.}}
    \label{fig:indo-adaptive-rag}
\end{figure*}

\subsection{Indonesian Language Adaptive RAG}
After obtaining the question classifier from the training and evaluation using the labeled data, the study progresses into the final part which is evaluating the Adaptive RAG system for the Indonesian language. In this stage, each input question is processed by the question classifier, which assesses its complexity and selects an appropriate answering strategy (non-retrieval, single-retrieval, or multi-retrieval). The selected strategy is then applied to generate the final answer. The system was evaluated using F1 score, exact match, accuracy, average time and average retrieval count. An overview of the entire system is illustrated in Figure \ref{fig:indo-adaptive-rag}.

\section{Experiments}
\label{sec:experiments}
In this section, we explained experimental setup, datasets, models, and implementation details for running every experiment in this study.

\subsection{Experimental Setup}
All experiments in this study, including machine translation, data labeling, fine-tuning the question classifier, and system evaluation were conducted on two identical machines. Each equipped with a Nvidia RTX 3090 GPU and 62 GB of RAM. While this hardware is sufficient for running LLM experiments additionally with sizable datasets, it also force us to use LLM that is relatively small and constraint the amount of LLM that can be utilized. Our study employs a search engine as the foundation of the retrieval process. Specifically, we use ElasticSearch, a widely adopted search and analytics engine designed for fast full-text search and efficient handling of large volumes of data in near real-time. To facilitate simplicity and rapid deployment, ElasticSearch is hosted using Docker. For the retrieval algorithm, we rely on BM25 \cite{bm25}, which serves as the default ranking function within the ElasticSearch system. The following section explains each answering method in detail, with an overview of the system illustrated in Figure \ref{fig:indo-adaptive-rag}.

\subsection{Datasets}
For this study, we utilize multiple question-answering datasets that have been shown to be of good quality. First, we use the IndoQA dataset \cite{indoqa}, which was published as part of a collaborative effort to develop datasets for Southeast Asian languages, focusing on natural language tasks across text, image, and auditory modalities \cite{seacrowd}. Second, we use the QASiNa dataset \cite{qasina}, which focuses on the religious domain for question-answering. This dataset was generated using a translation-based approach derived from Indo-SQuAD \cite{indo-squad}. Finally, we utilize the HotpotQA dataset that has been translated into Indonesian language using OPUS-MT translation model, specifically opus-mt-en-id, a neural machine translation model designed to translate text from English to Indonesian language. Additionally, we processed all rows in each dataset to ensure that no entries contained missing question-answer pairs. The details of each dataset are presented in Table \ref{table:dataset_info}.

\begin{table}[h!]
\caption{Dataset distribution based on partition.}
\label{table:dataset_info}
\centering
\begin{tabular}{|l|l|r|}
\hline
\textbf{Dataset Name} & \textbf{Partition} & \textbf{Amount of Data} \\
\hline
IndoQA & Training & 3,309 \\
IndoQA & Test     & 1,104 \\
QASina & All      & 500  \\
HotpotQA & Training     & 90,400\\
HotpotQA & Test     & 7,410 \\
\hline
\end{tabular}
\label{tab:dataset_distribution}
\end{table}

\subsection{Models}

Process of fine-tuning the question classifier, we utilized multiple variations of IndoBERT \cite{indobert}, a BERT based model trained using large Indonesian base language corpus and available through the Hugging Face model repository. The variations differ in model size and maximum sequence length (128 and 512). Details of the model used for the question classifier are presented in Table \ref{table:indobert_models}. The question classifier was fine-tuned for 10 epochs with an effective batch size of 16 (per-device batch size of 8 with gradient accumulation of 2). We used a learning rate of $2 \times 10^{-5}$, weight decay of 0.3, and 500 warmup steps. Dropout was applied with probabilities of 0.1 for both hidden layers and attention. Training and evaluation were performed at the end of each epoch, and gradient clipping was applied with a maximum norm of 1.

For the Adaptive RAG system, the LLMs used in the reasoning process are Gemma 3-4B \cite{gemma} and Qwen 3-8B \cite{qwen}, which are open-source large language models developed by Google and Alibaba, respectively. These models was selected not only for their state-of-the-art performance but also for their relatively low computational cost compared with other LLMs. For instance, LLaMA, despite having a comparable number of parameters, required significantly longer response times. Both Gemma 3 and Qwen 3 were hosted using Ollama, an open source framework designed to run LLMs efficiently on local machines.

\subsection{Implementation Details}
The automatic translation of the HotpotQA dataset was applied to both the training and testing dataset, Due to the large size of the dataset, the translation process required approximately one week to complete. For the labeling process, each instance from the IndoQA and QASiNa datasets was classified using the labeling pipeline explain earlier. In contrast, all questions in the HotpotQA dataset were directly assigned to label "C". Given the limited computational resources available to our study, we restricted the amount of data used in the Adaptive RAG evaluation phase. Specifically, we utilized 740 samples from the HotpotQA test set, 975 samples from the IndoQA test set, and the full QASiNa dataset, which consists of 500 samples.

\section{Results}
\label{sec:results}
In this section we are going to explain and showcase every experiment result, covering four key components: the translation of HotpotQA dataset, outcomes of labeling process result, the performance of the question classifier, and the evaluation of the Adaptive RAG system.

\subsection{HotpotQA Dataset Translation}
\begin{figure}[h]
    \centering
    \includegraphics[width=0.8\linewidth]{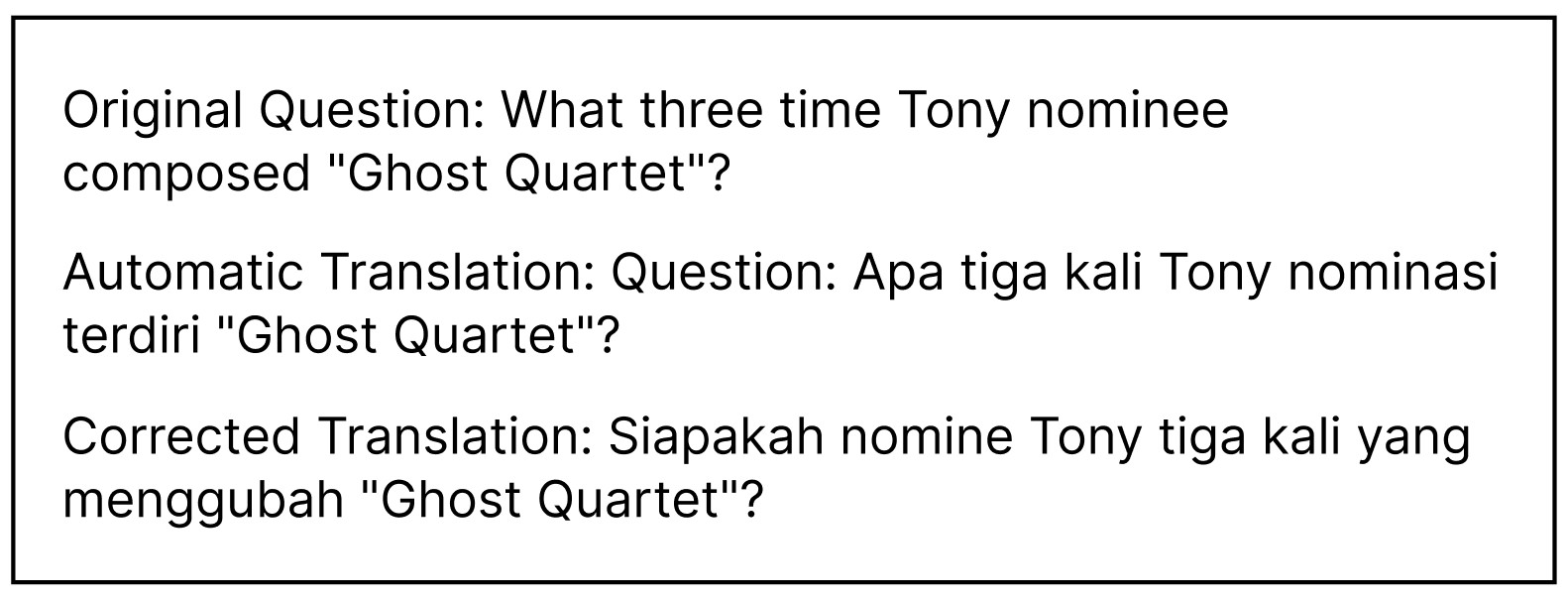}
    \caption{\textbf{Example of imperfect translation in the HotpotQA dataset.}}
    \label{fig:poor_translation}
\end{figure}

The automatic translation of the HotpotQA dataset produced a high-quality dataset, though not a perfect one. Based on limited manual evaluation, some deviation from the original meaning of the sentences remains. Figure \ref{fig:poor_translation} illustrates an example of an imperfect translation from the HotpotQA dataset.

The automatic translation of the HotpotQA dataset produces a generally high-quality dataset. However, the result were not all perfect. A manual inspection of the dataset revealed an occasional deviation from the original meaning of the sentences, which could introduce noise in the system. For example, as illustrated in Figure \ref{fig:poor_translation}, the translation of question ‘What three time Tony nominee composed "Ghost Quartet"?’ was translated as ‘Apa tiga kali Tony nominasi terdiri “Ghost Quartet”?’. This translation doesn't necessarily fail in converting each English word into Indonesian language, but it fails to preserve the grammatical structure and semantic meaning of the original sentence, as it incorrectly conveys "three times Tony nomination" instead of "three-time Tony nominee".

Despite these imperfections, the overall impact on our study is limited for two reasons. First, such mistranslations were relative rare compared to the total dataset size, meaning that the majority of data remains semantically correct. Second, our focus is on evaluating the Adaptive RAG system under realistic conditions, which is low-resource language setting, where noisy or imperfect data is an inherent challenge. Therefore, even though translation noise may introduce minor errors, it also makes the evaluation more similar to real-world deployment scenarios.

\subsection{Labeling Process Result}

\begin{figure*}
    \centering
    \includegraphics[width=.9\textwidth]{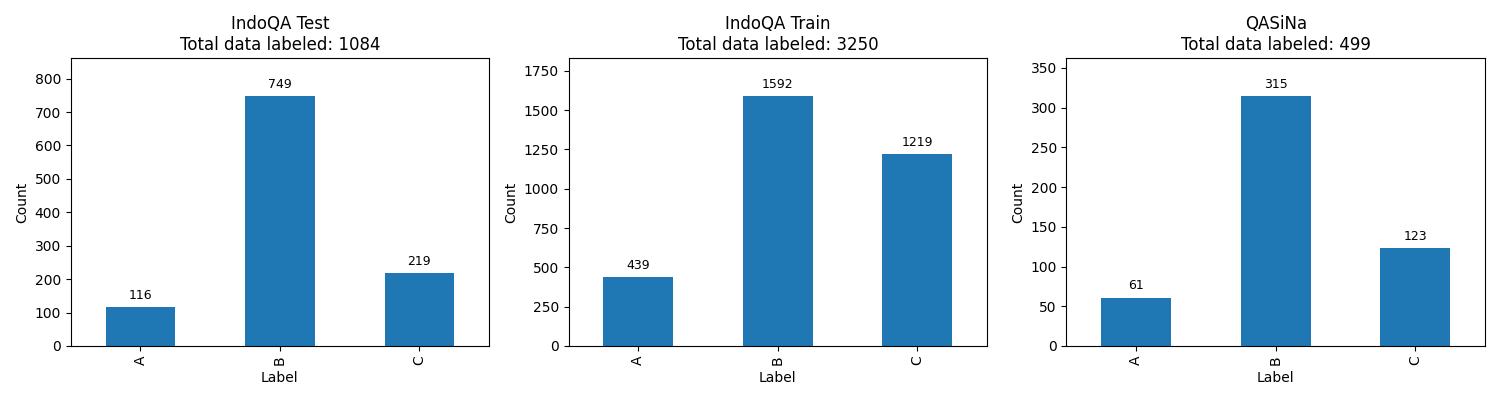}
    \caption{\textbf{Dataset Class Distribution after labeling process.}}
    \label{fig:label_distribution}
\end{figure*}

The labeling process assigned each question to one of three complexity categories ("A", "B", or "C"). Figure \ref{fig:label_distribution} illustrates the label distribution for IndoQA and QASiNa dataset. As expected, both of the dataset contained a higher proportion of "B" questions since the dataset was not necessarily created for the sake of multi-retrieval answering. It can also be seen that the distribution for class "A" is consistently the lowest, indicating that the parametric knowledge on which the LLM was trained on is insufficient to answer these questions correctly.

\subsection{Question Classifier}
The question classifier was trained to categorize questions into three different complexity levels. Enabling the Adaptive RAG system to dynamically select the most efficient answer strategy. From the results in Table \ref{table:indobert_models}, we can observe that all IndoBERT variations we used achieve consistent performance, with accuracy ranging between 0.724 - 0.756. Among the models IndoBERT Large P1 achieve the best overall performance with the lowest loss (1.651), the highest accuracy (0.756), and the strongest F1 score (0.760). This suggests a larger model capacity combined with shorted sequence length provides a better balance for classifying the questions we have. On the other hand, the performance difference between IndoBERT Base and Large is relatively small, showing that even the smaller model size can still perform against larger model for question complexity classification. 

\begin{figure}
    \centering
    \includegraphics[width=0.8\linewidth]{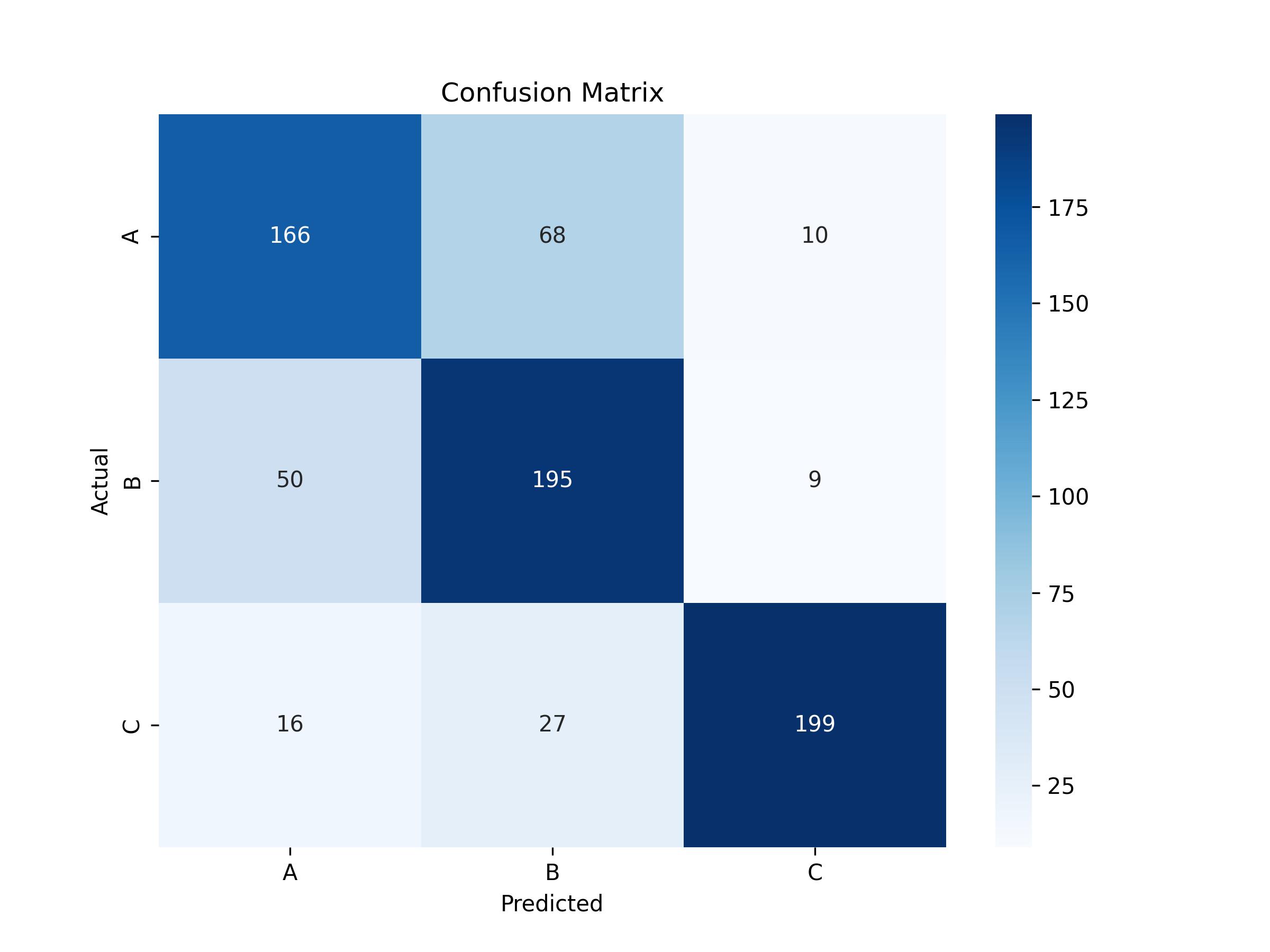}
    \caption{\textbf{IndoBERT Large P1 Confusion Matrix}}
    \label{fig:large_p1_cm}
\end{figure}

The confusion matrix of the IndoBERT Large P1 model is presented in Figure \ref{fig:large_p1_cm}. As shown, class "C" achieves the highest number of correct predictions, while classes "A" and "B" show a noticeable overlap, indicating room for improvement in distinguishing between them. Nevertheless, it is more important that the question classifier performs well on class "C", since questions in this category typically require further retrieval processes. Ensuring high accuracy for class "C" reduces the risk of misclassification and helps maintain the reliability of the overall system. To conclude the results, it shows our labeling process is reliable for creating a dataset that can be used for training a language model for recognizing the patterns of question complexity.

\begin{table}[h!]
\centering
\caption{Performance of IndoBERT variants on question classification.}
\label{table:indobert_models}
\begin{tabular}{|l|c|c|c|c|c|}
\hline
\textbf{Model} & \textbf{Loss} & \textbf{Accuracy} & \textbf{Precision} & \textbf{Recall} & \textbf{F1} \\
\hline
Base P1  & 1.658 & 0.727 & 0.736 & 0.728 & 0.731 \\
Base P2  & 1.718 & 0.724 & 0.729 & 0.725 & 0.727 \\
Large P1 & 1.651 & 0.756 & 0.766 & 0.756 & 0.760 \\
Large P2 & 1.665 & 0.745 & 0.758 & 0.746 & 0.750 \\
\hline
\end{tabular}
\end{table}

\subsection{Indonesian Language Adaptive RAG System}
This subsection presents the evaluation of the Adaptive RAG system applied to the Indonesian language. The detailed performance of the system is reported across three datasets: IndoQA, QASiNa, and HotpotQA (Tables \ref{tab:indo_qa_performance}, \ref{tab:qasina_performance}, and \ref{tab:hotpotqa_performance}). To provide a detailed analysis, the results are discussed per dataset.

\subsubsection{IndoQA Results}
On the IndoQA dataset, retrieval-based methods clearly shows a big performance different compared to non-retrieval method, confirming the importance of providing external information or non parametric knowledge to the LLM's query for question-answering. Single retrieval substantially improves the performance for both Gemma 3 and Qwen 3 model, reaching F1 score of 0.258 and 0.265 respectively. However, the impact multi-retrieval method proves unstable: while it slightly improves Qwen 3, it drastically reduces Gemma 3's performance compared to the single-retrieval method, this issue is explained in more detail later.

The adaptive retrieval strategy demonstrates strong potential for improving the system, even though its results are not higher than the single retrieval method. While the accuracy and F1 score remain close to those of single retrieval, adaptive retrieval achieves a lower step mean across both models. This indicates the classifier is effective at deciding when the LLM requires external information to answer a question. This highlights the overall performance is limited by the multi-retrieval method issues, which negatively affects the adaptive approach.

\subsubsection{QASiNa Results}
On the QASiNa dataset, the single-retrieval method again delivers the strongest performance across both Gemma 3 and Qwen 3, achieving the highest accuracy and F1 score (0.361 for Gemma 3 and 0.411 for Qwen 3). In contrast, the multi-retrieval method shows notably lower performance despite requiring substantially higher computational time, making it impractical for this task. Consistent with the IndoQA results, the adaptive retrieval strategy demonstrates promising potential, achieving F1 score close to those of single retrieval while maintaining a lower step mean, indicating its effectiveness in reducing unnecessary retrieval steps.

\subsubsection{HotpotQA Results}
On the HotpotQA dataset, the overall performance across all methods and model is noticeably poor, with accuracy and F1 score remaining below 0.06 for both Gemma 3 and Qwen 3. Non-retrieval and single-retrieval achieve slightly better results with exact match values around 40, but these remain very low in absolute terms. Multi-retrieval, however, performs especially poorly, yielding extremely low accuracy and F1 despite requiring very high computational time. Since the HotpotQA dataset is specifically designed for multi-hop reasoning, the adaptive strategy almost always selects the multi-retrieval method. As the result, adaptive retrieval performance mirrors the weaknesses of multi-retrieval, showing both very low scores (e.g., 0.008 accuracy and 0.008 F1 with Gemma 3) and excessive runtime costs. This correlation highlights a limitation of the adaptive approach: when the dataset primarily requires multi-retrieval, the system inherits the low performance of that method, indicating the need for improving multi-retrieval before adaptive strategies can be effective on complex datasets like HotpotQA.

\begin{table*}[htbp]
  \footnotesize
  \caption{Performance on the \textbf{IndoQA} Dataset (Total Data: 975)}
  \label{tab:indo_qa_performance}
  \centering
  \begin{tabular}{@{}p{3cm}p{2cm}c c c c c@{}}
    \toprule
    \textbf{Method} & \textbf{LLM Model} & \textbf{Exact Match} & \textbf{Accuracy} & \textbf{F1 Mean} & \textbf{Step Mean} & \textbf{Time Mean} \\
    \midrule
    Non Retrieval & Gemma 3 & 13 & 0.013 & 0.013 & 0.000 & 2.124 \\
    Single Retrieval & Gemma 3 & 252 & 0.258 & 0.258 & 1.000 & 4.187 \\
    Multi Retrieval & Gemma 3 & 18 & 0.018 & 0.019 & 2.826 & 127.681 \\
    \midrule
    Adaptive Retrieval & Gemma 3 & 206 & 0.211 & 0.213 & 1.012 & 17.175 \\
    \midrule
    Non Retrieval & Qwen 3 & 17 & 0.017 & 0.017 & 0.000 & 3.610 \\
    Single Retrieval & Qwen 3 & 257 & 0.264 & 0.265 & 1.000 & 12.215 \\
    Multi Retrieval & Qwen 3 & 91 & 0.093 & 0.094 & 0.005 & 26.050 \\
    \midrule
    Adaptive Retrieval & Qwen 3 & 200 & 0.205 & 0.206 & 0.661 & 11.134 \\
    \bottomrule
  \end{tabular}
\end{table*}

\begin{table*}[htbp]
  \footnotesize
  \caption{Performance on the \textbf{QASiNa} Dataset (Total Data: 500)}
  \label{tab:qasina_performance}
  \centering
  \begin{tabular}{@{}p{3cm}p{2cm}c c c c c@{}}
    \toprule
    \textbf{Method} & \textbf{LLM Model} & \textbf{Exact Match} & \textbf{Accuracy} & \textbf{F1 Mean} & \textbf{Step Mean} & \textbf{Time Mean} \\
    \midrule
    Non Retrieval & Gemma 3 & 9 & 0.018 & 0.018 & 0.000 & 1.663 \\
    Single Retrieval & Gemma 3 & 180 & 0.361 & 0.361 & 1.000 & 9.415 \\
    Multi Retrieval & Gemma 3 & 37 & 0.074 & 0.074 & 4.593 & 179.281 \\
    \midrule
    Adaptive Retrieval & Gemma 3 & 147 & 0.295 & 0.295 & 0.964 & 14.794 \\
    \midrule
    Non Retrieval & Qwen 3 & 13 & 0.026 & 0.026 & 0.000 & 2.878 \\
    Single Retrieval & Qwen 3 & 205 & 0.411 & 0.411 & 1.000 & 19.824 \\
    Multi Retrieval & Qwen 3 & 59 & 0.118 & 0.118 & 0.030 & 45.012 \\
    \midrule
    Adaptive Retrieval & Qwen 3 & 159 & 0.319 & 0.319 & 0.725 & 19.291 \\
    \bottomrule
  \end{tabular}
\end{table*}

\begin{table*}[htbp]
  \footnotesize
  \caption{Performance on the \textbf{HotpotQA} Dataset (Total Data: 740)}
  \label{tab:qasina_performance}
  \centering
  \begin{tabular}{@{}p{3cm}p{2cm}c c c c c@{}}
    \toprule
    \textbf{Method} & \textbf{LLM Model} & \textbf{Exact Match} & \textbf{Accuracy} & \textbf{F1 Mean} & \textbf{Step Mean} & \textbf{Time Mean} \\
     \midrule
    Non Retrieval & Gemma 3 & 43 & 0.058 & 0.059 & 0.000 & 0.670 \\
    Single Retrieval & Gemma 3 & 41 & 0.055 & 0.055 & 1.000 & 0.657 \\
    Multi Retrieval & Gemma 3 & 5 & 0.007 & 0.007 & 4.814 & 143.562 \\
    \midrule
    Adaptive Retrieval & Gemma 3 & 6 & 0.008 & 0.008 & 4.588 & 170.030 \\
    \midrule
    Non Retrieval & Qwen 3 & 38 & 0.051 & 0.051 & 0.000 & 2.785 \\
    Single Retrieval & Qwen 3 & 38 & 0.051 & 0.051 & 1.000 & 2.634 \\
    Multi Retrieval & Qwen 3 & 3 & 0.004 & 0.004 & 0.001 & 21.972 \\
    \midrule
    Adaptive Retrieval & Qwen 3 & 0 & 0.000 & 0.000 & 0.118 & 80.265 \\
    \bottomrule
  \end{tabular}
\end{table*}

\subsubsection{Multi Retrieval Issues}
The Multi-Retrieval method we employ follows the IRCot approach, which utilizes prompting to guide the retrieval system in obtaining relevant information. However, we observe a big drawback in the Gemma 3 and Qwen 3 reasoning ability when being used in Indonesian language, these of course largely impacted by both LLM's training data which is not dominant in Indonesian language. In our setup, multi-retrieval requires concatenating multiple documents into a single prompt which increases input length substrantially compared to single-retrieval answering method. As a result, both LLMs tend to hallucinate more frequently during the reasoning process; therefore, the guidance of the retrieval stage is impacted and ultimately reducing overall performance.

In practice, we observe that the multi-retrieval setup often fails to utilize the provided information effectively. For instance, given the question from HotpotQA dataset:

“Genealogi adalah supergroup Armenia yang bersaing dalam kontes menyanyi 2015 di kota apa?”, a translated question from "Genealogy is an Armenian supergroup that competed in a 2015 singing contest in what city?".

The system retrieves multiple contexts containing details about various singing contests, Armenia’s participation history, and related background information. However, the Gemma 3 model concludes with “Informasi tidak cukup untuk menjawab pertanyaan,” which is "Information is not enough to answer the question" failing to extract or reformulate any useful information for subsequent reasoning. This illustrates a common limitation of multi-retrieval: as input length grows due to concatenating several documents, the model’s reasoning quality degrades, often leading to unhelpful outputs or hallucinated answers.

In contrast, Gemini 2.5 Flash Lite, a larger model, handles the same prompt more effectively by producing focused keywords (“Genealogi”, “Armenia”, “kontes menyanyi 2015”). Rather than prematurely concluding that no answer is possible, it provides salient search terms that can be used in the next retrieval cycle. This behavior demonstrates an important capability for multi-retrieval: maintaining attention to core entities and events, while filtering out noise from longer prompts. It also highlights why larger models may mitigate—but not fully eliminate—the challenges of multi-retrieval in low-resource languages like Indonesian.

\section{Conclusion}
\label{sec:conclusion}
In this study, we attempted to bridge the language barrier for LLMs, especially the Adaptive RAG system which demonstrate a tremendous performance compared to other method that doesn't keep in mind the LLM's parametric knowledge or question's complexity over computational and time cost \cite{adaptive_rag}. The system incorporates a question complexity classifier which was trained using a dataset that has gone through a labeling process. We also did automated translation of a multi-hop specific question answering dataset, because there is no dataset in Indonesian language for that specific task. 

Question complexity classifier enables the system to adaptively select the most suitable answering strategy. Three methods were defined: (1) non-retrieval, where the LLM answers only utilizing its parametric knowledge; (2) single-retrieval, which augments the LLM with one external retrieval step; and (3) multi-retrieval, which follows the Interleaving Retrieval with Chain-of-Thought Reasoning approach to address complex questions by guiding information retrieval steps through the LLM’s reasoning process. This adaptive design enhances the system’s flexibility in handling questions of varying complexity.

The outcomes of our study demonstrate a strong potential of applying Adaptive RAG system in low-resource settings, particularly for the Indonesian language. This is supported by the accuracy and reliability of our question complexity classifier, which ensures that the answering strategy selected by the system is dependable. However, a key limitation of our research is the performance of the multi-retrieval answering method. Its effectiveness is constrained by LLM's reasoning capabilities, as the model frequently hallucinates when guiding the information retrieval process. Consequently, when evaluated on a multi-hop question answering dataset, overall performance declined significantly due to the shortcomings of the multi-retrieval approach.

\textbf{Future Improvements} — Given the poor performance of the multi-retrieval approach, we recommend continuing research on question answering with a stronger focus on multi-hop question answering tasks. Several directions for improvement include:

\begin{itemize}
    \item Developing a dataset that is originally written in Indonesian language rather than translated from English, in order to avoid issues of mistranslation and meaning drift.
    \item Utilizing or training an LLM that is specifically pre-trained or fine-tuned on Indonesian language, so it can better understand the context of multi-retrieval prompts and guide the information retrieval process more effectively.
    \item Improving the multi-retrieval method itself, or even designing a retrieval approach tailored for low-resource languages, thereby reducing the need to rely solely on fine-tuning large models.
\end{itemize}

\section{Acknowledgment}
We would like to send our deepest gratitude towards Bina Nusantara University for supporting us in this particular study.

\end{document}